\RequirePackage{fix-cm}
\documentclass[smallextended]{svjour3}       
\smartqed  
\usepackage{graphicx}

\usepackage{subcaption}
\usepackage{multirow}
\usepackage{float}
\usepackage{amsmath}
 \journalname{Cognitive Computation}
\begin{document}

\title{A Deep Decoder Structure Based on Word Embedding Regression for An Encoder-Decoder Based Model for Image Captioning
}
\subtitle{}


\author{Ahmad Asadi         \and
        Reza Safabakhsh 
}


\institute{Ahmad Asadi \at
              Department of Computer Engineering and Information Technology, Amirkabir University of Technology, Hafez St., Tehran, Iran \\
              \email{ahmad.asadi@aut.ac.ir}           
           \and
           Reza Safabakhsh \at
              Department of Computer Engineering and Information Technology, Amirkabir University of Technology, Hafez St., Tehran, Iran\\
              \email{safa@aut.ac.ir}       
}

\date{Received: date / Accepted: date}

\maketitle

\begin{abstract}

Generating textual descriptions for images has been an attractive problem for the computer vision and natural language processing researchers in recent years. Dozens of models based on deep learning have been proposed to solve this problem. The existing approaches are based on neural encoder-decoder structures equipped with the attention mechanism. These methods strive to train decoders to minimize the log likelihood of the next word in a sentence given the previous ones, which results in the sparsity of the output space. In this work, we propose a new approach to train decoders to regress the word embedding of the next word with respect to the previous ones instead of minimizing the log likelihood. The proposed method is able to learn and extract long-term information and can generate longer fine-grained captions without introducing any external memory cell. Furthermore, decoders trained by the proposed technique can take the importance of the generated words into the consideration while generating captions. In addition, a novel semantic attention mechanism is proposed that guides attention points through the image, taking the meaning of the previously generated word into account. We evaluate the proposed approach with the MS-COCO dataset. The proposed model outperformed the state of the art models especially in generating longer captions. It achieved a CIDEr score equal to 125.0 and a BLEU-4 score equal to 50.5, while the best scores of the state of the art models are 117.1 and 48.0, respectively.

\end{abstract}

\section{Introduction}

Image captioning is the task of generating a textual description for a given image. The generated description is required to be syntactically and grammatically correct and it should suggest a holistic explanation of the image. Image caption generator models can be widely used for providing advanced conceptual search tools on images \cite{fang2015captions}, assisting blind people \cite{wu2017automatic}, and human-robot interactions \cite{das2017visual}.

The problem of generating an appropriate caption for an image is a challenging problem in the computer vision and natural language processing fields. Most methods using deep neural networks for solving this task are based on an encoder-decoder baseline inspired by sequence-to-sequence methods in machine translation \cite{sutskever2014sequence}. 

The encoder employs convolutional neural networks(CNNs) for understanding the image concept, details, containing objects, and object relations. In fact, the encoder makes a translation from the input space to a latent space, and extracts a feature vector from the given image to be used by a decoder later. The decoder uses recurrent neural networks(RNNs) for generating appropriate sentences in a word by word manner, describing the information and features extracted from the image.

Coping with the long-term dependencies while generating the caption is one of the most important challenges in the simple encoder-decoder based models. These models strive to extract a single feature vector which restricts the encoder to encode all the necessary information in a fixed length vector. In addition, the decoder has to extract all the required information from that single fixed length vector while generating the words of the caption.

Attention mechanism solved the problem of the simple encoder-decoder baseline in the machine translation task \cite{bahdanau2014neural}. The visual attention mechanism proposed by Mnih et al. \cite{mnih2014recurrent} applied the idea to the image caption generation task. The keypoint of the attention mechanism solution is to generate more than one feature vector for image representation in the encoding phase and select and use the appropriate feature vectors for word generation in the decoding phase. This mechanism enables models to attend to salient parts of an image while generating its caption.

The visual attention mechanism has recently been widely used in the proposed deep image caption generating models. CNN extracts information from the spatial segments of an image and encodes the extracted information into multiple feature vectors. The employed RNN then uses one of the extracted feature vectors at each step, while creating the words of the caption. It is similar to the process of attending to a specific area of an image while telling a word about that specific part and going to the next area for the next word till the whole caption is generated.

Designing a decoder structure able to generate rich fine-grained sentences plays an important role in improving the performance of image captioning models. Typical encoder-decoder based models for image captioning train the decoder using the log likelihood of each word in the sentence given the previously generated words. Minimizing the log likelihood function requires "one-hot" vectors for both the decoder output and the desired label at each step. Using the one-hot encoding strongly increases the sparsity of the output space of the model which makes the training phase harder. In this work, instead of using loglikelihood minimization to train the decoder, we utilize word embedding regression which resolves the sparsity of the output space, decreases the number of the decoder's weights, and allows training deeper structures to extract more complex features and finer-grained captions.

Our work makes the following contributions: 1) We propose a novel method to train decoders to regress the embedding of each word in the sentence instead of predicting the probability distribution of the words at each step. 2) The proposed model resolves the problem of sparsity of the output space and results in faster convergence of the decoder. 3) We introduce a novel attention mechanism that takes the meaning of the word generated in the last step into account. 4) The proposed method outperforms the state-of-the-art on image captioning benchmarks.

The rest of this paper is organized as follows: In section \ref{related} we review related models based on the encoder-decoder framework in the image captioning field. In section \ref{method} the main problems of vanishing gradients and output space sparsity in stacked decoders are presented and a novel technique is introduced to cope with these problems. The proposed model is then evaluated and the evaluation results are reported and discussed in section \ref{results}. Finally, the conclusions of the paper are summarized in section \ref{conc}.
\section{Related Work}\label{related}

Recent studies in image caption generation can be categorized into two major categories: the alignment methods and the encoder-decoder based methods. The studies in each category are reviewed in this section. 

\subsection{Alignment Methods}

The general approach of the studies in this group is to make an alignment between images and their corresponding captions. All methods in this category achieve this using the following general pipeline architecture:

\begin{enumerate}
	\item A CNN is applied on the input image to extract a good feature vector as the image representation. In fact, the CNN makes a transformation from the input space to a $d$-dimensional latent space.
	\item Simultaneously an RNN is used to extract a good representation of the corresponding caption. The only constraint is that the caption's extracted representation should be in the same $d$-dimensional space as the image latent space.
	\item In the training phase, the CNN and the RNN are jointly trained in a way that the similarity of the image and its corresponding representation is maximized, while the similarity of the image and other captions' representations is minimized.
	\item In the testing phase, the input image is fed into the trained CNN to generate an appropriate representation. the generated representation then is used to generate or retrieve the desired caption. Furthermore, if the input is a sentence, the trained RNN is used to generate the appropriate representation and the image with the most similar representation is retrieved as the search result.
\end{enumerate}

\begin{figure}
	\centering
	\includegraphics[scale=0.6]{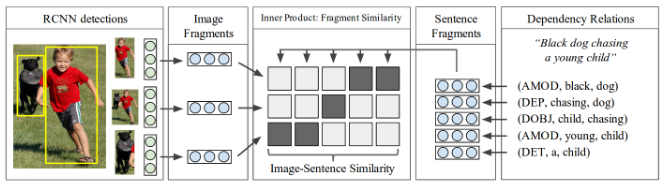}
	\caption{Structure of alignment method proposed by Karpathy et al.\cite{karpathy2014deep}}
	\label{fig:k2014}
\end{figure}

Approaches to image-caption alignment differ with respect to the structure of the used CNNs and RNNs. Karpathy et al.\cite{karpathy2014deep} proposed a deep bidirectional alignment between images and their captions. In this work, the input image is first fragmented to 19 sub-spaces using the R-CNN method \cite{girshick2014rich}. In the next step, a CNN \cite{krizhevsky2012imagenet} is applied to the whole image and the 19 extracted sub-spaces and a 4096-dimensional feature vector is extracted for the 20 images. Meanwhile, a dependency tree of a given caption in the training set is extracted and its relationships are used to identify sentence fragments. A simple linear transformation is applied to each sentence fragment in order to generate a 4096-dimensional meaning vector for all of the extracted sentence fragments. An alignment model is trained in order to maximize the similarity of the related parts of the image and the fragments of the sentence. Figure \ref{fig:k2014} demonstrates the system architecture\cite{karpathy2014deep}.

Karpathy et al. also used a bidirectional RNN as a word embedding model and trained it with captions provided in the training set \cite{karpathy2015deep}.

\subsection{Encoder-Decoder Based Methods}

The encoder-decoder framework proposed by Cho et al.\cite{cho2014learning} is one of the most popular models used in machine translation. Methods based on this framework split the translation task into two steps: 1) extract features from the source sentence which is called the \textit{encoding phase}, 2) generate a new sentence in the destination language given the meaning encoded into the feature vector, which is called the \textit{decoding phase}. 

Since, the encoder-decoder framework yields an end-to-end model to solve sophisticated problems, it is employed as a solution to a wide variety of problems in computer vision and natural language processing fields. Venugopalan et al. \cite{venugopalan2014translating} first employed the encoder-decoder model in video description generation. Sun et al. \cite{sun2018emotional} also proposed a model for emotional human-machine conversation generation using the encoder-decoder baseline.

Image captioning task is one of the main application areas of the end-to-end neural encoder-decoder based models. Xu et al. first proposed a model based on this framework in image caption generation, substituting the source sentence in machine translation with the input image \cite{xu2015show}. Thus, the \textit{encoder} changed in a way that it generates a feature vector given an image and the remaining parts of the framework remained unchanged.

Encoder-decoder based image captioning baseline model has been employed by many researchers to propose novel image description systems. Wang et al. used a CNN with two novel bidirectional recurrent neural networks to model complicated linguistic patterns using historical and future sentence context\cite{wang2016image}. Vinyals et al. used the \textit{"Inception"} model proposed by Szegedy et al.\cite{szegedy2016rethinking} as the encoder and LSTMs as the decoder of its framework\cite{vinyals2015show}. In addition, a linear transformation layer was added at the input of the LSTMs for better training.

More sophisticated models based on the encoder-decoder framework were also proposed. Ding et al. \cite{ding2018neural} proposed a method based on the encoder-decoder framework, called \textit{Reference based Long Short Term Memory} (R-LSTM), aiming to lead the model to generate a more descriptive sentence for a given image by introducing some reference information. In this work, by introducing reference information, the importance of different words is considered while generating captions.

The first idea of using visual attention was proposed by Mnih et al.\cite{mnih2014recurrent} to cope with the problem of fixed length context vectors extracted by the encoders. In this work, in order to decrease the overhead of applying convolutional networks at each step, a sequence of feature vectors was extracted from different image regions. At the first step, a convolutional network was applied to all selected image regions and the extracted feature vectors, called annotation vectors, were listed. At each step of generating text, a new feature vector was calculated from the given annotation vectors that helped to predict the next word in the sentence.

Attention mechanism then was exported to machine translation by Bahdanau et al. \cite{bahdanau2014neural}. Furthermore, Ba et al.\cite{ba2014multiple} employed the attention mechanism for multiple object recognition. Finally, the well-formed encoder-decoder empowered by attention with two different mechanisms called \textit{"soft attention"} and \textit{"hard attention"} was proposed by Xu et al.\cite{xu2015show}.

Advances made through deploying the attention based techniques encouraged researchers to focus on this kind of model. A large number of studies tried to improve the simple attention model in image captioning. Park et al.\cite{park2017attend} employed an encoder-decoder method along with a memory slut to generate a personalized caption for an image in social media for specific users. Cornia et al.\cite{cornia2017visual} used saliency map estimators to strengthen the attention mechanism. Yang et al.\cite{yang2016stacked} used the attention mechanism in visual question answering in order to find the best place to attend in image while generating the answer. In this work, attention layers were structured in a stack. This improved the model's performance. Furthermore, Donahue et al.\cite{donahue2015long} proposed an encoder-decoder framework for video description using the attention mechanism.

More complex models were proposed employing the attention mechanism to improve the quality of generated captions. Chen et al. \cite{chen2017temporal} proposed a novel method to use the attention mechanism and an RNN in order to first \textit{observe} the input image and create different weights for each word of its caption while training to learn better from the key information of the caption. Chen et al. \cite{chen2018show} also proposed a technique to focus on training a good attribute-inference model via an RNN and the attention mechanism where the  co-occurrence  dependencies  among  attributes are maintained. Chen et al. \cite{chen2019image} proposed a memory-enhanced captioning model for image captioning to cope with the problem of long-term dependencies. In this work, an external memory is introduced to keep the information of all of the generated words at previous steps. Using the introduced external memory, RNN cells can make better prediction at each step without trying to extract the information from the hidden state. 

To the best of our knowledge, decoders in all of the previous studies are used to model the log-likelihood of the next word, given the previously generated words and the image context vector according to expression \eqref{eq:loglikelihood}. In this expression $W$ is the set of trainable parameters, $y_i$ is the generated word at step $i$ and $\theta$ is the image context vector.

\begin{equation}
-\Sigma_{t=0}^L \>\> log \> P_W(y_t| y_{t-1}, y_{t-2}, \cdots, y_0, \theta)
\label{eq:loglikelihood}
\end{equation}

In order to compute the probability distribution of expression \eqref{eq:loglikelihood}, the last layer of the decoder should have the same size as that of the vocabulary dictionary.

The expectation of each component of the output vector is extremely small, resulting in extremely small back-propagated gradients of the errors. For the typical image captioning datasets, the expected value of gradient of errors is in the range of $5e^{-5}$. Small gradients of error make learning more difficult and decrease the model's convergence speed. This is the most important obstacle to make deeper decoders for memorizing long-term dependencies.

Gu et al. proposed a stacked decoder architecture for image caption generation \cite{gu2018stack}. In this research, the problem of vanishing gradients in training deep stacked decoders is addressed by providing a learning objective function which enforces intermediate supervisions.

In this work, a new method is proposed to cope with the problems of modeling log-likelihood. The proposed method changes the optimization problem used to train the decoder weights in a way that the decoder performs a regression over the word embeddings instead of predicting their probability distributions. In this way, not only the sparsity of the output space is resolved, but also higher gradients are produced at the last layer of the decoder and the gradients are less likely to vanish. In this way, we can use deeper stacked decoders which in general extract more complex features and are able to generate longer qualified captions. In addition, a novel attention mechanism is proposed to take the meaning of the last generated word into account, while creating the visual attention point at each step.

\section{Proposed Method}\label{method}

We propose a novel approach to train decoders for image captioning. The main idea is to use a word-embedding vector instead of a one-hot vector representation as the decoder desired output. As a result, the optimization problem changes from predicting the conditional probability distribution of the next word to a word-embedding regression. In addition, a new attention mechanism is proposed to take the last word meaning into consideration. 

\subsection{Typical Decoder}

A typical decoder in image captioning task is an RNN whose parameters are found by solving the optimization problem \eqref{eq:opt1}, in which $W$ denotes the set of trainable weights, $y_t$ denotes the generated word at time step $t$, $\theta$ denotes the representation vector of the input image, and $N_d$ denotes the number of words in the generated sentence.

\begin{equation}
minimize \>\>\>\> - \Sigma_{t=0}^{N_d} \>\> log \> P_W ( y_t| y_{t-1}, \cdots, y_0, \theta) 
\label{eq:opt1}
\end{equation}

In order to model the probability distribution, a softmax layer is used at the end of the decoder to normalize the output logits according to \eqref{eq:softmax} in which $S_i$ denotes the $i$th component of the predicted probability distribution, $f_i$ denotes the $i$th element of the output logits and $C$ is the output size.

\begin{equation}
S_i = \frac{e^{f_i}}{\Sigma_{i=0}^Ce^{f_i}} \label{eq:softmax}
\end{equation}

Typically, the softmax cross entropy loss function $L = -log(S_y)$ is used in image captioning models. The derivative of the loss function with respect to its weights $W_i$ can be computed using the chain rule as shown in \eqref{eq:chFin}.

\begin{align}
\frac{\delta L}{\delta W_i} &= \frac{\delta L}{\delta S_y} \frac{\delta S_y}{\delta f_i} \frac{\delta f_i}{\delta W_i}\\
\frac{\delta L}{\delta S_y} &= \frac{\delta (- log(S_y))}{\delta S_y} = - \frac{1}{S_y} \\
\frac{\delta S_y}{\delta f_i} &= 
\begin{cases}
S_y(1-S_i) &\quad\text{i = y}\\
-S_iS_y &\quad\text{i} \neq \text{y} \\
\end{cases}\\	
\frac{\delta L}{\delta W_i} &= 
\begin{cases}
(S_i-1)\frac{\delta f_i}{\delta W_i}  &\quad\text{i = y}\\
S_i\frac{\delta f_i}{\delta W_i}  &\quad\text{i = y}\\
\end{cases} \label{eq:chFin}
\end{align}

A typical language model should use about 20,000 words for sentence generation (based on the size of current image captioning benchmarks). This means $ E(S_i) \approx \frac{1}{20000} = 5e^{-5}$. Therefore, the average value of $\Delta W_i$ is of order $5e^-5 \frac{\delta f_i}{\delta W_i}$. The small value of the back-propagated error is also multiplied by a small learning rate at each layer and gets even smaller. This means that the backpropagated error makes no significant change in the weights of the first layers of the model.

\subsection{Proposed Decoder}

Making the decoder deeper results in adding more non-linearity to the model. This is shown to be helpful in vision tasks \cite{simonyan2014very}. Stacking LSTM layers on top of each other improves the decoder performance if the back-propagated error is large enough to make changes in weight values and train them. 

We used the word-embedding vector as the desired output of the decoder at each step instead of the one-hot vector. In order to use the word-embedding vector, the optimization problem \eqref{eq:opt1} should be changed. The problem is changed from predicting the probability distribution of each word to the regression of the word-embedding of the new word. The new optimization problem is formulated as in \eqref{eq:opt2} in which $\Gamma(y_t)$ is the embedding vector of the word $y_t$ and $f_t$ is the output logits of the decoder at time step $t$.

\begin{equation}
minimize \>\>\>\> ||f_t - \Gamma(y_t)||_2
\label{eq:opt2}
\end{equation}

\begin{figure}[h]
	\centering
	\includegraphics[scale=0.7]{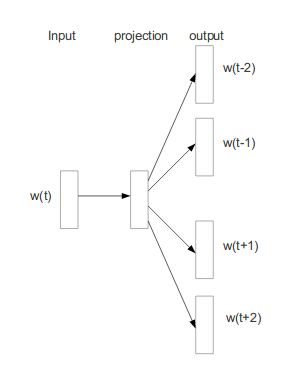}
	\caption{An illustration of the skip-gram word embedding model proposed in \cite{mikolov2013efficient}}
	\label{fig:embd}
\end{figure}

The word embedding model used in this work is the model proposed by Mikolov et al.\cite{mikolov2013efficient}. The \textit{skip-gram} model is selected as it can significantly facilitate the training of the decoder. As figure \ref{fig:embd} shows, in the skip-gram model, the embedding is trained in a way that the prediction error of the previous and the next word embeddings is minimized. This means that the embedding of a word gives hints to predict the word embedding of the next ones. So, estimating the next word probability distribution will be easier for the decoder.

Using word embedding as the input vector of all LSTM cells results in a notable decrease in the dimensionality of the input space and therefore, decreases the number of parameters (network weights) of the model. Thus, in this case, we can use a stacked multi-RNN cell in the decoder block. 

Using a stacked multi-RNN cell structure empowers the model to predict longer captions. In a stack of LSTMs, remembering the far away words will become more convenient because each LSTM cell can be imagined as a cell responsible to remember a far word or context in a sentence.

\subsection{Encoder}

In this work, the Inception V3 network proposed by Szegedy et al. \cite{szegedy2016rethinking} is used as the encoder because of its good performance on image classification and feature extraction tasks. The outputs of one of the layers of the Inception V3 which is called \textit{transfer learning layer} are used as "annotation vector" and passed to the decoder.

\subsection{Proposed Attention Mechanism}

We denote the annotation vector set extracted by the encoder (the output of the transfer learning layer of the Inception V3 model) as 
$H = \{ h_i | 0 \leq i \leq N_d - 1\}$. 

At time step $t$, while generating the $t$th word of the caption, the system computes the used context according to equations \eqref{eq:att1} to \eqref{eq:att3}. In equation \eqref{eq:att1}, $c_t$ denotes the context vector and $\alpha_{tj}$ the weight of the $j$th annotation vector at step $t$. A \textit{SoftMax} layer is used to compute each attention coefficient $\alpha_{tj}$ according to equation \eqref{eq:att2}. In this equation, $e_{tj}$ is a measure of how good the annotation vector $h_j$ is for generating a feature vector at step $t$. $e_{tj}$ is computed according to equation \eqref{eq:att3} in which $s_{t-1}$ denotes the hidden state of the decoder at step $t-1$, $\nu_{t-1}$ is the embedding vector of the word predicted at step $t-1$, and $f$ is implemented by a multilayer perceptron network with a single hidden layer. The weights of this MLP are trained jointly with the weights of the decoder during the training phase.

\begin{align}
&c_t = \Sigma_{j=1}^L \alpha_{tj}h_j
\label{eq:att1}
\\
&\alpha_{tj} = \frac{exp(e_{tj})}{\Sigma_{k=1}^Lexp(e_{tk})}
\label{eq:att2}
\\
&e_{tj} = f(s_{t-1}, h_j, \nu_{t-1})
\label{eq:att3}
\end{align}

The above mentioned attention mechanism is the general structure represented by Xu et al.\cite{xu2015show}, except that the similarity function $f$ takes an extra input parameter $\nu_{t-1}$ to guide the attention considering the previous predicted word.

Function $f$ is implemented with an MLP represented with equation \eqref{eq:attmlp}. In this equation, $N_d$ gives the number of annotation vectors, $k$ denotes the number of MLP's hidden layer neurons, $b_1$ and $b_2$ are bias vectors of the first and second layers of the network, and all $W_{ij}$s denote the weight vectors of the network.

\begin{equation}
f(s_{t-1}, h_j, \nu_{t-1}) = W_{N_dk}.(W_{kh}.h_j + W_{k\nu}.\nu_{t-1} + b_1) + W_{N_ds}s_{t-1} + b_2
\label{eq:attmlp}
\end{equation}

\subsection{Regularizations}

\begin{figure}[h]
	\centering
	\includegraphics[scale=0.12]{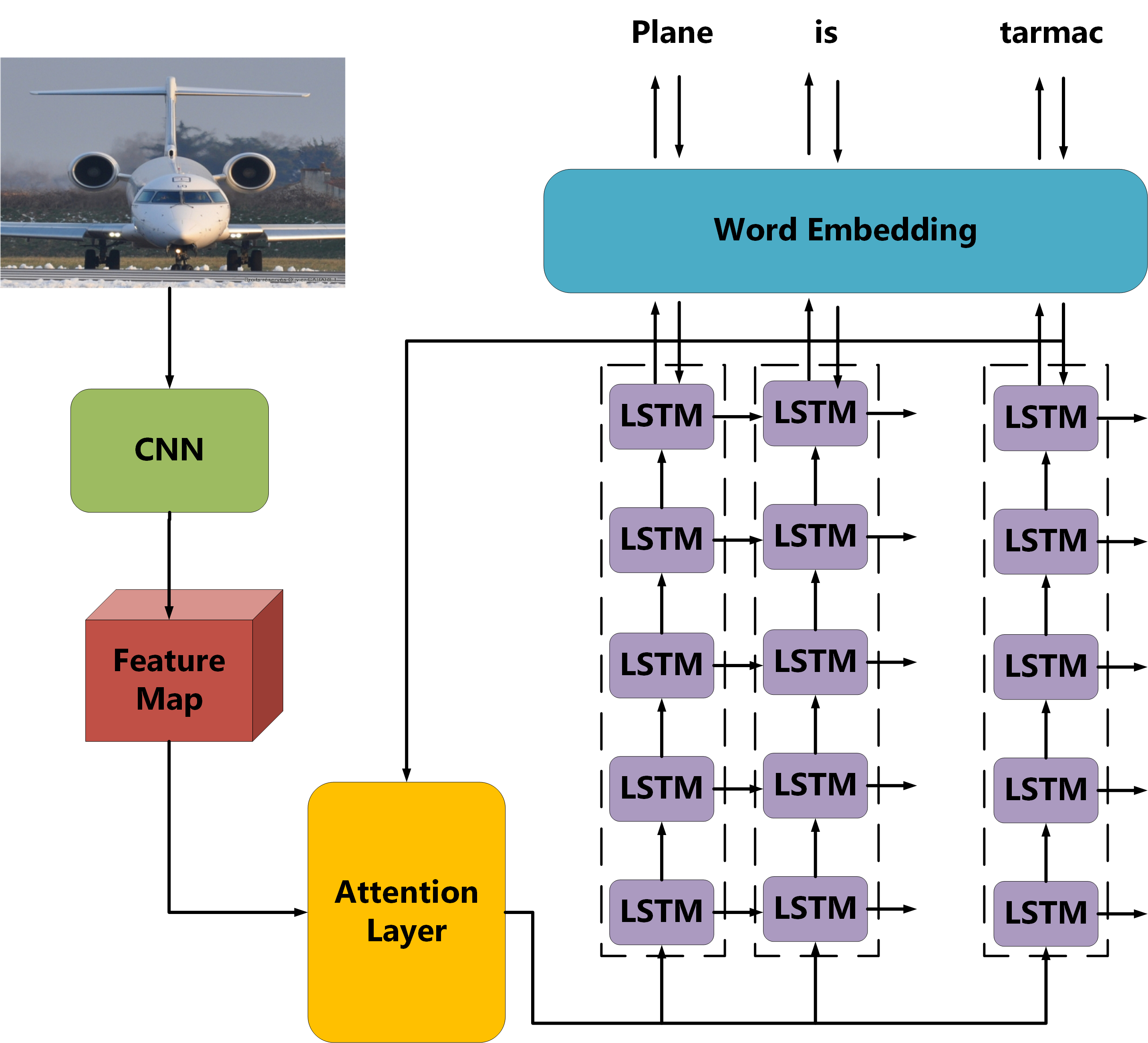}
	\caption{The proposed architecture for image caption generation task.}
	\label{fig:whole}
\end{figure}

In order to prevent overfitting on the training dataset, the dropout technique proposed by Srivastava et al.\cite{srivastava2014dropout} is employed and a dropout-wrapper is used on all of the units in the proposed network. In addition, The output of the transfer learning layer of the Inception module is pyramided down into a quarter of its original size to decrease the number of required parameters in the attention layer. Figure \ref{fig:whole} demonstrates the whole structure of proposed method. In order to make a more readable figure, direct input word vectors at each time step are removed from figure.

\section{Experimental Results}\label{results}

The experiments carried out to evaluate the performance of the proposed method are explained in this section. Section 4.1 takes a brief look at the available dataset and evaluation metrics used to evaluate the model performance. The implementation details are explained in section 4.2. Section 4.3 reports the results obtained from these experiments and presents some discussions regarding the results.

\subsection{Dataset and Evaluation Metrics}

The MSCOCO dataset intoduced by Lin et al.\cite{lin2014microsoft} is used to evaluate the model proposed in this work. The training set of MSCOCO consists of more than 82,789 images, each with at most five human generated captions. The validation and testing sets of this dataset have 40,504 and 40,000 images respectively, captioned in the same way. 

The proposed method is evaluated based on the popular image captioning evaluation metrics BLEU score \cite{papineni2002bleu}, CIDEr score\cite{vedantam2015cider}, ROUGE\_L score \cite{lin2004rouge} and METEOR score \cite{banerjee2005meteor}. 

\subsection{Implementation Details}

Each input image is fed to a convolutional network to extract high level features. We used \textit{Inception V3} \cite{szegedy2016rethinking} as the encoder in this work. The output of the \textit{Transfer Layer} of the \textit{Inception} model are used as the annotation vectors of the given image. \textit{Transfer Layer} is the last layer before the fully connected block of the \textit{Inception} model. The output of this layer shapes a set of feature vectors each associated with a specific spatial segment of the source image. The used \textit{Inception} model is pretrained on the \textit{ImageNet} dataset \cite{deng2009imagenet}. Since the feature maps generated by this model are general enough to extract the necessary information for sentence generation, no additional training is required. Annotation vectors extracted by the \textit{Inception} model are sub-sampled one level using an extra max-pooling layer to reduce the weights of the attention layer model. The annotation vectors are fed to the stacked multi-RNN cell through the attention layer in the next step.

The "\textit{Gensim}" word2vec model proposed by Mikolov et al.\cite{mikolov2013efficient} is employed for word embedding in this work. The word-embedding model is trained on all captions of the training set once before all other processes. At each step in caption generation, the embedding vector of the previously generated word is fed to the attention mechanism and the stacked multi-RNN cell as input. Furthermore, all LSTM cells are trained to predict the embedding vector of the next word instead of one-hot vector. This is done using the mean squared error as the cost function and the Adam optimizer. All of our evaluations use a model with 8-layer stacked multi-RNN cells trained with a 1024-dimensional word embedding model and an initial learning rate of 0.001 and a decay rate of 0.1 at each step.

\subsection{Results}

In this section the experimental methodology and quantitative and qualitative results are described. The effectiveness of the proposed model for image captioning is also validated. In the following experiments, we denote the simple stacked architecture without attention layer with "\textit{stacked}". Also we used "\textit{ATT}" for the stacked architecture with a single attention layer only before the first layer. The numbers in parentheses specifies the number of layers in the stacked decoder.

\subsubsection{Comparison with the state-of-the-art models}

The proposed method is evaluated and compared with other existing models on the MS-COCO dataset. The results are reported in table \ref{tbl:res1}. For the existing methods, both recent best performing methods and the baseline models are chosen and their results are directly taken from the existing literature. The table columns present scores for the metrics BLEU-4 (B4) to BLEU-1 (B1), METEOR (M), CIDEr(C) and ROUGE-L (R).  

\begin{table}[H]
	\centering
	\caption{Performance of the proposed method compared to the state-of-the-art methods on MS-COCO dataset.}
	\label{tbl:res1}
	\begin{tabular}{| c | c | c | c | c | c | c | c |}
		\hline
		& R		&M		&C		&B4		&B3		&B2		&B1
		\\
		\hline
		Vinyals et al. 2017\cite{vinyals2017show}		& 		53.0 		&		25.4		&		94.3		&		30.9 		&		40.7		&		54.2		&		71.3
		\\
		Lu et al. 2017\cite{lu2017knowing} 		& 		55.0 		&		26.4		&		104.0		&		33.6 		&		44.4		&		58.4 		&		74.8
		\\
		Chen et al. 2017\cite{chen2017temporal} & 54.7 & 25.9 & 105.9 & 32.4 & 44.1 & 59.1 & 75.7
		\\
		Ren  et al. 2017\cite{ren2017deep} & 52.5 & 25.1 & 93.7 & 30.4 & 40.3 & 53.9 & 71.3
		\\
		Gu et al. 2017\cite{gu2017empirical} & - & 25.1 & 99.1 & 30.4 & 40.9 & 54.6 & 72.1
		\\
		Rennie et al. 2017\cite{rennie2017self} & 56.3 & 27.0 & 114.7 & 35.2 & - & - & -
		\\
		Wang et al. 2017\cite{wang2017skeleton} & 48.9 & 24.7 & 96.6 & 25.9 & 35.5 & 48.9 & 67.3
		\\
		Gan et al. 2017\cite{gan2017semantic} & 54.3 & 25.7 & 100.3 & 34.1 & 44.4 & 57.8 & 74.1
		\\
		Liu et al. 2018\cite{liu2018show} & 57.0 & 27.4 & 117.1 & 35.8 & 48.0 & 63.1 & 80.1
		\\		
		Wang et al. 2018\cite{wang2018image} & - & 21.1 & 69.5 & 23.0 & 33.3 & 47.4 & 65.6
		\\
		Cornia et al. 2018\cite{cornia2018paying} & 52.1 & 24.8 & 89.8 & 28.4 & 39.1 & 53.6 & 70.8
		\\
		Ding et al. 2018\cite{ding2018neural} & 55.5 & 26.1 & 105.5 & 34.2 & 45.8 & 60.5 & 76.8
		\\
		Chen et al. 2018\cite{chen2018show} & 54.9 & 33.8 & 104.4 & 33.8 & 44.3 & 57.9 & 74.3
		\\
		Gu et al. 2018\cite{gu2018stack} & - & 27.4 & 120.4 & 36.1 & 47.9 & 62.5 & 78.6
		\\
		Chen et al. 2019\cite{chen2019image} & 58.7 & 28.7 & \textbf{125.5} & 38.4 & 50.7 & 65.5 & \textbf{81.9}
\\
\hline
		Stacked + ATT (8)
		& 
		\textbf{64.9} 
		&
		\textbf{34.7}
		&
		125.0
		&
		\textbf{50.5}
		&
		\textbf{57.1}
		&
		\textbf{66.4}
		&
		73.7
		\\
		\hline
	\end{tabular}
\end{table}

According to the table \ref{tbl:res1}, except for BLEU@1 the performance of the method proposed in this paper is better than or roughly equal to those of the reference models. In addition, without introducing any external memory cell, our model is generating relatively better captions than the model proposed by Chen et al. \cite{chen2019image}. This shows that making deeper stacked decoder structures empowers the RNNs to memorize longer history of the generated words' information. The proposed method specially outperforms the existing models based on traditional RNNs (without any external memory cell) with respect to BLEU-4 and CIDEr factors measuring the similarity of the same 4-grams and the longest weighted sub-sequence of the same words in suggested and reference captions, respectively. This means our model can generate captions similar to reference human generated image descriptions that are longer than those generated by other methods. It shows that substituting the typical prediction problem at each step to generate caption words with a regression problem resolves the problem of long-term dependencies in the encoder-decoder based models.

\subsubsection{On the depth of the decoder}

The proposed model is trained with different number of stacked decoder layers. Table \ref{tbl:layers} reports the number of the trainable parameters of the decoder with respect to the depth of the decoder when using the one-hot vector representation versus when using the word embedding. 

\begin{table}[h]
	\centering
	\caption{The number of decoder's trainable parameters with different decoder depths for two representation methods.}
	\label{tbl:layers}
	\begin{tabular}{|c|c c c c c|}
		\hline
	\multirow{2}{7em}{\centering Representation Methods}
		&\multicolumn{5}{|c|}{Decoder Depth}
\\
& 1& 2& 5& 8 &10
\\
		\hline
		Embedding& 17.8 M & 26.2 M & 51.4 M & 76.6 M & 93.4 M \\ 
		One-Hot& 37.38 M & 60.2 M & 118.21 M & 189.8 M & 233.5 M \\ 
		\hline
		
	\end{tabular}
\end{table}

In addition, table \ref{tbl:res2} reports the experimental results of training the stacked decoder structure with different depths. According to this table, we can now stack up to 8 layers of LSTMs on top of each other and use them as the decoder. This means that in our model, the gradients produced at the last layer of the decoder can be back propagated to the first layer without vanishing, while the other models with stacked decoders can have up to only 2 layers \cite{gu2018stack} \cite{donahue2015long}.

\begin{table}[h]
	\centering
	\caption{Performance of different decoders with different depths compared to each other.}
	\label{tbl:res2}
	\begin{tabular}{| c |c |c |c |c |c |c |c | }
		\hline
		Model & R & M & C & B4 & B3 & B2 & B1 
		\\
		\hline
		Stacked(5) & 62.7 & 33.1 & 122.7 & 46.3 & 52.7 & 65.9 & 71.3
		\\
		Stacked(8)
		& 
		\textbf{64.9} 
		&
		\textbf{34.7}
		&
		\textbf{125.0}
		&
		\textbf{50.5}
		&
		\textbf{57.1}
		&
		\textbf{66.4}
		&
		\textbf{73.7}
		\\
		Stacked(10) & 60.8 & 31.3 & 115.7 & 40.0 & 49.8 & 60.9 & 70.5
		\\
		\hline		
	\end{tabular}
\end{table}

\subsubsection{Qualitative Results}

\begin{figure}[h]
	\centering
	\begin{subfigure}{0.45\textwidth}
		\centering
		\includegraphics[scale=0.25]{cor1.jpg}
		\caption{
			\footnotesize 
			LSTM:
			Giraffe standing in a tree filled area\\
			GT: 
			A giraffe standing next to a forest filled with trees.\\
			A giraffe eating food from the top of the tree.\\
		}
	\end{subfigure}
	\hfil
	\begin{subfigure}{0.45\textwidth}
		\centering
		\includegraphics[scale=0.25]{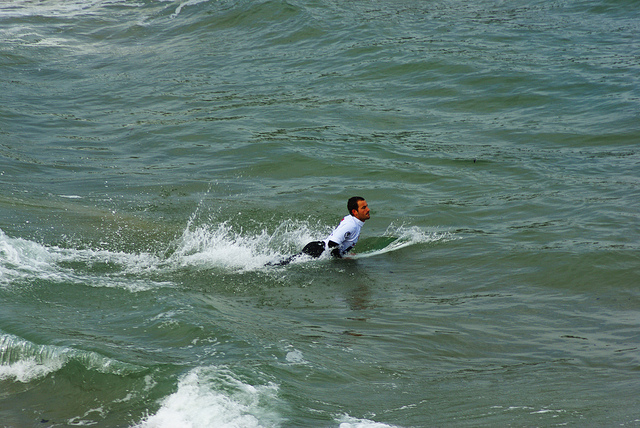}
		\caption{
			\footnotesize			
			LSTM: Man on a surf board in the ocean\\
			GT: A‌ man laying on a surfboard in the water.\\
			A man lying on a surfboard in some small waves in water.\\
		}
	\end{subfigure}
	\\
	\begin{subfigure}{0.45\textwidth}
		\centering
		\includegraphics[scale=0.25]{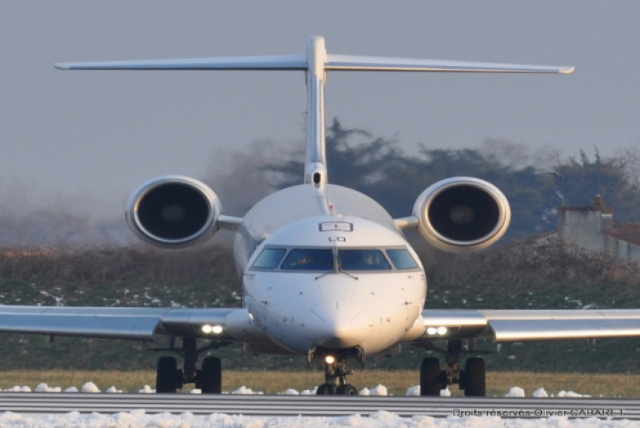}
		\caption{
			\footnotesize			
			LSTM: Plane is on the ground on the tarmac\\
			GT: A modern jet airliner on a snow edged runway.\\
			The airplane is on the ground on the runway.\\
		}
	\end{subfigure}
	\caption{Samples of correct generated captions}
	\label{tbl:corr}
\end{figure}

\begin{figure}[h]
	\centering
	\begin{subfigure}{0.45\textwidth}
		\centering
		\includegraphics[scale=0.25]{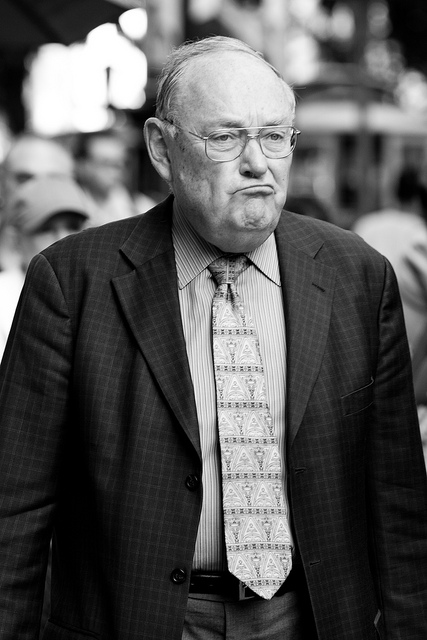}
		\caption{
			\footnotesize 
			LSTM:
			Old man in a coat has a giraffis look on his face\\
			GT: 
			A man stands with a frown on his face.\\
			A old man in coat and tie walking down a busy street.\\
		}
	\end{subfigure}
	\hfil
	\begin{subfigure}{0.45\textwidth}
		\centering
		\includegraphics[scale=0.25]{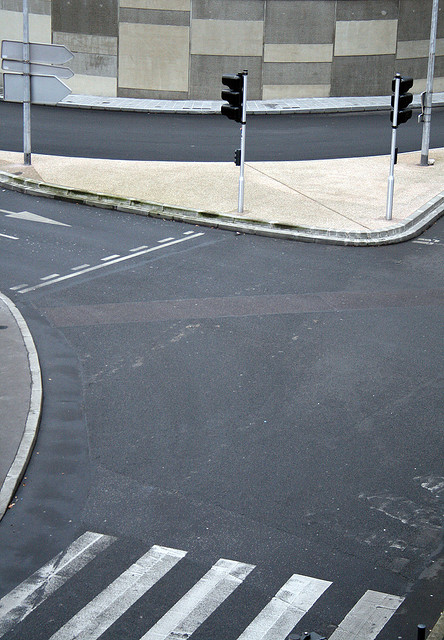}
		\caption{
			\footnotesize			
			LSTM: Train of an empty intersection with traffic lights\\
			GT: Two traffic lights are posted near the street intersection.\\
			An empty street with some stop lights in a little island.\\
		}
	\end{subfigure}
	\\
	\begin{subfigure}{0.45\textwidth}
		\centering
		\includegraphics[scale=0.25]{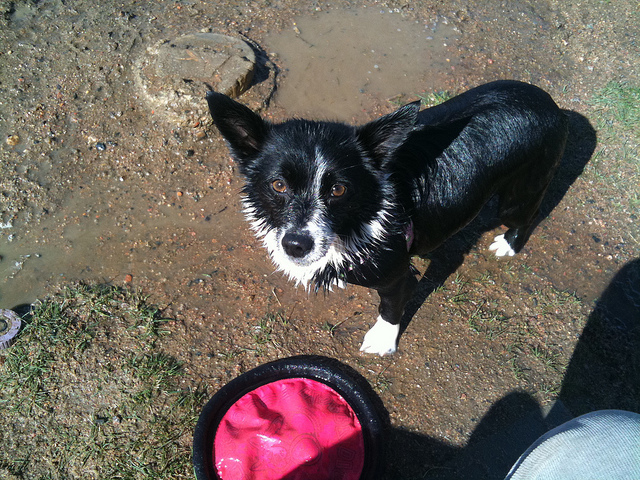}
		\caption{
			\footnotesize			
			LSTM: small black and dog with a frisbee by its feet\\
			GT: A small dog standing on a wet ground looking up.\\
			A small black and white dog standing on a sparse grass looking at ahuman.\\
		}
	\end{subfigure}
	\caption{Samples of incorrect generated captions}
	\label{tbl:incorr}
\end{figure}

Figures \ref{tbl:corr} and \ref{tbl:incorr} display some samples of correct and incorrect captions generated by the proposed method.
In addition, our method outperforms the existing methods using METEOR and ROUGE-L measures. This means better words are used in captions generated by our work. This is the result of training the LSTM cells to predict the embedding of the next words instead of their one-hot vectors. Indeed, it allows the decoder to use alternative synonym words more easily than before at the word generation phase and takes the word co-occurrences and meanings into account while choosing them in the caption generation phase.

\section{Conclusions} \label{conc}

In this paper, we proposed a novel approach to train the decoders in encoder-decoder structures for the image captioning task. We used a word-embedding vector as the desired output of the decoder instead of a one-hot vector. This enables the stacked decoders to cope with the long-term dependencies and vanishing gradients. In addition, this allows the decoders to be deeper and have more layers stacked on top of each other. Furthermore, we introduced a new attention mechanism to gain better order of attention points during caption generation. We used previously generated words as extra inputs to the attention mechanism and trained a neural network to find the next focus point on the image using the meaning of the previously predicted word and the current state vector of the decoder. This network is trained jointly with the other parts of the decoder during the training phase. In addition, we employed a stacked multi-RNN cell as the decoder and trained it to predict the embedding of the next word instead of its one-hot vector in order to first take word meanings into consideration and second reduce model parameters while generating captions for given images. Results show that the proposed method outperforms the existing models with respect to measures corresponding to length of the caption on the most widely used dataset of image captioning.

\newpage
  \bibliographystyle{ieeetr} 
\bibliography{ref}   

\begin{thebibliography}{10}

\bibitem{fang2015captions}
H.~Fang, S.~Gupta, F.~Iandola, R.~K. Srivastava, L.~Deng, P.~Doll{\'a}r,
  J.~Gao, X.~He, M.~Mitchell, J.~C. Platt, {\em et~al.}, ``From captions to
  visual concepts and back,'' in {\em Proceedings of the IEEE conference on
  computer vision and pattern recognition}, pp.~1473--1482, 2015.

\bibitem{wu2017automatic}
S.~Wu, J.~Wieland, O.~Farivar, and J.~Schiller, ``Automatic alt-text:
  Computer-generated image descriptions for blind users on a social network
  service.,'' in {\em CSCW}, pp.~1180--1192, 2017.

\bibitem{das2017visual}
A.~Das, S.~Kottur, K.~Gupta, A.~Singh, D.~Yadav, J.~M. Moura, D.~Parikh, and
  D.~Batra, ``Visual dialog,'' in {\em Proceedings of the IEEE Conference on
  Computer Vision and Pattern Recognition}, vol.~2, 2017.

\bibitem{sutskever2014sequence}
I.~Sutskever, O.~Vinyals, and Q.~V. Le, ``Sequence to sequence learning with
  neural networks,'' in {\em Advances in neural information processing
  systems}, pp.~3104--3112, 2014.

\bibitem{bahdanau2014neural}
D.~Bahdanau, K.~Cho, and Y.~Bengio, ``Neural machine translation by jointly
  learning to align and translate,'' {\em arXiv preprint arXiv:1409.0473},
  2014.

\bibitem{mnih2014recurrent}
V.~Mnih, N.~Heess, A.~Graves, {\em et~al.}, ``Recurrent models of visual
  attention,'' in {\em Advances in neural information processing systems},
  pp.~2204--2212, 2014.

\bibitem{karpathy2014deep}
A.~Karpathy, A.~Joulin, and L.~F. Fei-Fei, ``Deep fragment embeddings for
  bidirectional image sentence mapping,'' in {\em Advances in neural
  information processing systems}, pp.~1889--1897, 2014.

\bibitem{girshick2014rich}
R.~Girshick, J.~Donahue, T.~Darrell, and J.~Malik, ``Rich feature hierarchies
  for accurate object detection and semantic segmentation,'' in {\em
  Proceedings of the IEEE conference on computer vision and pattern
  recognition}, pp.~580--587, 2014.

\bibitem{krizhevsky2012imagenet}
A.~Krizhevsky, I.~Sutskever, and G.~E. Hinton, ``Imagenet classification with
  deep convolutional neural networks,'' in {\em Advances in neural information
  processing systems}, pp.~1097--1105, 2012.

\bibitem{karpathy2015deep}
A.~Karpathy and L.~Fei-Fei, ``Deep visual-semantic alignments for generating
  image descriptions,'' in {\em Proceedings of the IEEE Conference on Computer
  Vision and Pattern Recognition}, pp.~3128--3137, 2015.

\bibitem{cho2014learning}
K.~Cho, B.~Van~Merri{\"e}nboer, C.~Gulcehre, D.~Bahdanau, F.~Bougares,
  H.~Schwenk, and Y.~Bengio, ``Learning phrase representations using rnn
  encoder-decoder for statistical machine translation,'' {\em arXiv preprint
  arXiv:1406.1078}, 2014.

\bibitem{venugopalan2014translating}
S.~Venugopalan, H.~Xu, J.~Donahue, M.~Rohrbach, R.~Mooney, and K.~Saenko,
  ``Translating videos to natural language using deep recurrent neural
  networks,'' {\em arXiv preprint arXiv:1412.4729}, 2014.

\bibitem{sun2018emotional}
X.~Sun, X.~Peng, and S.~Ding, ``Emotional human-machine conversation generation
  based on long short-term memory,'' {\em Cognitive Computation}, vol.~10,
  no.~3, pp.~389--397, 2018.

\bibitem{xu2015show}
K.~Xu, J.~Ba, R.~Kiros, K.~Cho, A.~Courville, R.~Salakhudinov, R.~Zemel, and
  Y.~Bengio, ``Show, attend and tell: Neural image caption generation with
  visual attention,'' in {\em International Conference on Machine Learning},
  pp.~2048--2057, 2015.

\bibitem{wang2016image}
C.~Wang, H.~Yang, C.~Bartz, and C.~Meinel, ``Image captioning with deep
  bidirectional lstms,'' in {\em Proceedings of the 2016 ACM on Multimedia
  Conference}, pp.~988--997, ACM, 2016.

\bibitem{szegedy2016rethinking}
C.~Szegedy, V.~Vanhoucke, S.~Ioffe, J.~Shlens, and Z.~Wojna, ``Rethinking the
  inception architecture for computer vision,'' in {\em Proceedings of the IEEE
  Conference on Computer Vision and Pattern Recognition}, pp.~2818--2826, 2016.

\bibitem{vinyals2015show}
O.~Vinyals, A.~Toshev, S.~Bengio, and D.~Erhan, ``Show and tell: A neural image
  caption generator,'' in {\em Proceedings of the IEEE conference on computer
  vision and pattern recognition}, pp.~3156--3164, 2015.

\bibitem{ding2018neural}
G.~Ding, M.~Chen, S.~Zhao, H.~Chen, J.~Han, and Q.~Liu, ``Neural image caption
  generation with weighted training and reference,'' {\em Cognitive
  Computation}, pp.~1--15, 2018.

\bibitem{ba2014multiple}
J.~Ba, V.~Mnih, and K.~Kavukcuoglu, ``Multiple object recognition with visual
  attention,'' {\em arXiv preprint arXiv:1412.7755}, 2014.

\bibitem{park2017attend}
C.~C. Park, B.~Kim, and G.~Kim, ``Attend to you: Personalized image captioning
  with context sequence memory networks,'' {\em arXiv preprint
  arXiv:1704.06485}, 2017.

\bibitem{cornia2017visual}
M.~Cornia, L.~Baraldi, G.~Serra, and R.~Cucchiara, ``Visual saliency for image
  captioning in new multimedia services,'' in {\em Multimedia \& Expo Workshops
  (ICMEW), 2017 IEEE International Conference on}, pp.~309--314, IEEE, 2017.

\bibitem{yang2016stacked}
Z.~Yang, X.~He, J.~Gao, L.~Deng, and A.~Smola, ``Stacked attention networks for
  image question answering,'' in {\em Proceedings of the IEEE Conference on
  Computer Vision and Pattern Recognition}, pp.~21--29, 2016.

\bibitem{donahue2015long}
J.~Donahue, L.~Anne~Hendricks, S.~Guadarrama, M.~Rohrbach, S.~Venugopalan,
  K.~Saenko, and T.~Darrell, ``Long-term recurrent convolutional networks for
  visual recognition and description,'' in {\em Proceedings of the IEEE
  conference on computer vision and pattern recognition}, pp.~2625--2634, 2015.

\bibitem{chen2017temporal}
H.~Chen, G.~Ding, S.~Zhao, and J.~Han, ``Temporal-difference learning with
  sampling baseline for image captioning,'' 2017.

\bibitem{chen2018show}
H.~Chen, G.~Ding, Z.~Lin, S.~Zhao, and J.~Han, ``Show, observe and tell:
  Attribute-driven attention model for image captioning.,'' in {\em IJCAI},
  pp.~606--612, 2018.

\bibitem{chen2019image}
H.~Chen, G.~Ding, Z.~Lin, Y.~Guo, C.~Shan, and J.~Han, ``Image captioning with
  memorized knowledge,'' {\em Cognitive Computation}, pp.~1--14, 2019.

\bibitem{gu2018stack}
J.~Gu, J.~Cai, G.~Wang, and T.~Chen, ``Stack-captioning: Coarse-to-fine
  learning for image captioning,'' in {\em Thirty-Second AAAI Conference on
  Artificial Intelligence}, 2018.

\bibitem{simonyan2014very}
K.~Simonyan and A.~Zisserman, ``Very deep convolutional networks for
  large-scale image recognition,'' {\em arXiv preprint arXiv:1409.1556}, 2014.

\bibitem{mikolov2013efficient}
T.~Mikolov, K.~Chen, G.~Corrado, and J.~Dean, ``Efficient estimation of word
  representations in vector space,'' {\em arXiv preprint arXiv:1301.3781},
  2013.

\bibitem{srivastava2014dropout}
N.~Srivastava, G.~Hinton, A.~Krizhevsky, I.~Sutskever, and R.~Salakhutdinov,
  ``Dropout: A simple way to prevent neural networks from overfitting,'' {\em
  The Journal of Machine Learning Research}, vol.~15, no.~1, pp.~1929--1958,
  2014.

\bibitem{lin2014microsoft}
T.-Y. Lin, M.~Maire, S.~Belongie, J.~Hays, P.~Perona, D.~Ramanan,
  P.~Doll{\'a}r, and C.~L. Zitnick, ``Microsoft coco: Common objects in
  context,'' in {\em European conference on computer vision}, pp.~740--755,
  Springer, 2014.

\bibitem{papineni2002bleu}
K.~Papineni, S.~Roukos, T.~Ward, and W.-J. Zhu, ``Bleu: a method for automatic
  evaluation of machine translation,'' in {\em Proceedings of the 40th annual
  meeting on association for computational linguistics}, pp.~311--318,
  Association for Computational Linguistics, 2002.

\bibitem{vedantam2015cider}
R.~Vedantam, C.~Lawrence~Zitnick, and D.~Parikh, ``Cider: Consensus-based image
  description evaluation,'' in {\em Proceedings of the IEEE conference on
  computer vision and pattern recognition}, pp.~4566--4575, 2015.

\bibitem{lin2004rouge}
C.-Y. Lin, ``Rouge: A package for automatic evaluation of summaries,'' {\em
  Text Summarization Branches Out}, 2004.

\bibitem{banerjee2005meteor}
S.~Banerjee and A.~Lavie, ``Meteor: An automatic metric for mt evaluation with
  improved correlation with human judgments,'' in {\em Proceedings of the acl
  workshop on intrinsic and extrinsic evaluation measures for machine
  translation and/or summarization}, pp.~65--72, 2005.

\bibitem{deng2009imagenet}
J.~Deng, W.~Dong, R.~Socher, L.-J. Li, K.~Li, and L.~Fei-Fei, ``Imagenet: A
  large-scale hierarchical image database,'' in {\em Computer Vision and
  Pattern Recognition, 2009. CVPR 2009. IEEE Conference on}, pp.~248--255,
  IEEE, 2009.

\bibitem{vinyals2017show}
O.~Vinyals, A.~Toshev, S.~Bengio, and D.~Erhan, ``Show and tell: Lessons
  learned from the 2015 mscoco image captioning challenge,'' {\em IEEE
  transactions on pattern analysis and machine intelligence}, vol.~39, no.~4,
  pp.~652--663, 2017.

\bibitem{lu2017knowing}
J.~Lu, C.~Xiong, D.~Parikh, and R.~Socher, ``Knowing when to look: Adaptive
  attention via a visual sentinel for image captioning,'' in {\em Proceedings
  of the IEEE Conference on Computer Vision and Pattern Recognition (CVPR)},
  vol.~6, 2017.

\bibitem{ren2017deep}
Z.~Ren, X.~Wang, N.~Zhang, X.~Lv, and L.-J. Li, ``Deep reinforcement
  learning-based image captioning with embedding reward,'' {\em arXiv preprint
  arXiv:1704.03899}, 2017.

\bibitem{gu2017empirical}
J.~Gu, G.~Wang, J.~Cai, and T.~Chen, ``An empirical study of language cnn for
  image captioning,'' in {\em Proceedings of the International Conference on
  Computer Vision (ICCV)}, 2017.

\bibitem{rennie2017self}
S.~J. Rennie, E.~Marcheret, Y.~Mroueh, J.~Ross, and V.~Goel, ``Self-critical
  sequence training for image captioning,'' in {\em CVPR}, vol.~1, p.~3, 2017.

\bibitem{wang2017skeleton}
Y.~Wang, Z.~Lin, X.~Shen, S.~Cohen, and G.~W. Cottrell, ``Skeleton key: Image
  captioning by skeleton-attribute decomposition,'' in {\em Proceedings of the
  IEEE Conference on Computer Vision and Pattern Recognition}, pp.~7272--7281,
  2017.

\bibitem{gan2017semantic}
Z.~Gan, C.~Gan, X.~He, Y.~Pu, K.~Tran, J.~Gao, L.~Carin, and L.~Deng,
  ``Semantic compositional networks for visual captioning,'' in {\em
  Proceedings of the IEEE Conference on Computer Vision and Pattern
  Recognition}, vol.~2, 2017.

\bibitem{liu2018show}
X.~Liu, H.~Li, J.~Shao, D.~Chen, and X.~Wang, ``Show, tell and discriminate:
  Image captioning by self-retrieval with partially labeled data,'' {\em arXiv
  preprint arXiv:1803.08314}, 2018.

\bibitem{wang2018image}
C.~Wang, H.~Yang, and C.~Meinel, ``Image captioning with deep bidirectional
  lstms and multi-task learning,'' {\em ACM Transactions on Multimedia
  Computing, Communications, and Applications (TOMM)}, vol.~14, no.~2s, p.~40,
  2018.

\bibitem{cornia2018paying}
M.~Cornia, L.~Baraldi, G.~Serra, and R.~Cucchiara, ``Paying more attention to
  saliency: Image captioning with saliency and context attention,'' {\em ACM
  Transactions on Multimedia Computing, Communications, and Applications
  (TOMM)}, vol.~14, no.~2, p.~48, 2018.

\end{thebibliography}


\end{document}